\documentclass[10pt,twocolumn,letterpaper]{article}

\usepackage{cvpr}      % To produce the REVIEW version
\definecolor{cvprblue}{rgb}{0.21,0.49,0.74}
\usepackage[pagebackref,breaklinks,colorlinks,allcolors=cvprblue]{hyperref}

\def\paperID{*****} % *** Enter the Paper ID here
\def\confName{CVPR}
\def\confYear{2025}

\title{ ContextRefine-CLIP  for EPIC-KITCHENS-100  Multi-Instance \\ Retrieval  Challenge 2025}

\author{Jing He, Yiqing Wang, Lingling Li, Kexin Zhang, Puhua Chen \\
Intelligent Perception and Image Understanding Lab, Xidian University \\
Xi’an, Shannxi Province, 710071, China\\
{\tt\small \{24171213874, 24171213882\}@stu.xidian.edu.cn}
}

\begin{document}
\maketitle
\begin{abstract}
This report presents ContextRefine-CLIP (CR-CLIP), an efficient model for visual-textual multi-instance retrieval tasks. The approach is based on the dual-encoder AVION, on which we introduce a cross-modal attention flow module to achieve bidirectional dynamic interaction and refinement between visual and textual features to generate more context-aware joint representations. For soft-label relevance matrices provided in tasks such as EPIC-KITCHENS-100, CR-CLIP can work with Symmetric Multi-Similarity Loss to achieve more accurate semantic alignment and optimization using the refined features. Without using ensemble learning, the CR-CLIP model achieves 66.78\% mAP and 82.08\% nDCG on the EPIC-KITCHENS-100 public leaderboard, which significantly outperforms the baseline model and fully validates its effectiveness in cross-modal retrieval. The code will be released open-source on \href{https://github.com/delCayr/ContextRefine-Clip}{https://github.com/delCayr/\href{https://github.com/delCayr/ContextRefine-Clip}{ContextRefine-Clip} }.
\end{abstract}    
\section{Introduction}
\label{sec:intro}

Visual-text retrieval, as a key bridge connecting visual content and natural language descriptions, plays an increasingly important role in content search, multimedia recommendation, human-computer interaction, and other fields. In recent years, the performance of visual-text retrieval has been significantly improved with the development of deep learning techniques\cite{li2019visual}, especially the emergence of CLIP models and their variants based on contrast learning \cite{wang2024variance}\cite{le2020contrastive}. These models have learned powerful cross-modal joint embedding representation capabilities by pre-training on large-scale graphic-text pairs.

However, in complex scenarios with multiple instances and soft labeling such as EPIC-KITCHENS-100\cite{damen2018scaling}\cite{damen2022rescaling}, existing methods still face challenges in fully exploiting such fine-grained correlation information. While advanced loss functions such as Symmetric Multi-Similarity Loss  (SMSLoss)\cite{wang2024sms} work to optimize the learning objective for such soft labels by more precisely defining the relationships between pairs of positive and negative samples, we observe that there is still room for further optimization at the feature interaction level of the model architecture. Most existing models, including model like AVION \cite{zhao2023training}that focus on efficient training, tend to achieve modal alignment only through simple projection operations after independent visual and text encoders. This approach, while efficient, may limit the model's ability to capture deep inter-modal dependencies and contextual dynamics, especially when subtle semantic differences need to be distinguished. When both visual content and textual descriptions contain rich and potentially ambiguous information, the lack of deep modal interactions may lead to the generation of joint representations that are not discriminative enough in the face of complex semantic correspondences.

The ContextRefine-CLIP proposed in this paper is able to go beyond the traditional global feature matching to more deeply model the fine-grained correspondence between specific actions and textual verbs in videos, and between combinations of objects and descriptive statements in complex scenes. The context-aware joint features generated by the cross-modal attention mechanism not only have stronger discriminative power, but also work more effectively with advanced loss functions such as SMSLoss to achieve more accurate semantic alignment and optimization under soft label supervision. 

In our experiments, without employing ensemble learning and relying solely on data augmentation, our single-model CR-CLIP achieves 1st place on the EPIC-KITCHENS-100 Multi-Instance Retrieval Challenge 2025 public leaderboard. Compared to the ensemble-based 2024 solution, which achieved an average mAP of 63.8\% and nDCG of 74.3\%, our approach improves the average mAP to 66.8\% (+3.0\%) and nDCG to 82.1\% (+7.8\%), demonstrating the strong effectiveness and robustness of our method in a more lightweight setting.

\section{Method}
\label{sec:formatting}

To utilize soft-label information more efficiently and enhance deep inter-modal interactions in visual-textual multi-instance retrieval tasks, we propose the ContextRefine-CLIP model. The model introduces a novel cross-modal Contextual Refinement module based on the efficient  AVION architecture to generate more discriminative and context-aware joint embedding representations.
%-------------------------------------------------------------------------
\subsection{Model Architecture}

The overall architecture of ContextRefine-CLIP follows a classic dual-transformer design, consisting of a Vision Transformer (ViT) for visual encoding and a standard Transformer for text encoding. A key component of the model is the Cross-Modal Context Refinement (CMCR) module, which refines and aligns visual and textual features through bidirectional attention before projection. The refined features are then mapped into a shared embedding space via separate MLP projection heads. The model is trained using the Symmetric Multi-Similarity  Loss to fully exploit soft label correlations. 

For input representation, video frames are sampled, augmented, patchified, and embedded for the visual encoder, while text is tokenized with special markers, embedded, and passed through the text encoder.
%-------------------------------------------------------------------------
\subsection{Cross-modal Context Refinement Module}

To enhance the depth and precision of multi-modal interaction prior to the final projection stage, we introduce a Cross-Modal Context Refinement (CMCR) module built upon a pair of lightweight cross-attention layers. This design enables bidirectional semantic refinement, where visual features are dynamically updated with textual context and vice versa, thereby producing context-aware, semantically aligned representations for downstream retrieval tasks. 

Our core component is Cross-Attention Layer, which is based on the standard mechanism of multi-attention and allows Q from one modality to interact with K and V provided by another modality. Given a visual feature vector $F_v \in \mathbb{R}^{B \times D_v}$ and a textual feature vector  $F_t \in \mathbb{R}^{B \times D_t}$, we project both into a common attention dimension $D_a$, as in Eq. (1), (2), (3).
\begin{equation}
Q_t = W_Q^t F_t
  \label{eq:important}
\end{equation}
\begin{equation}
 K_v = W_K^v F_v
  \label{ eq:important}
 \end{equation}
\begin{equation}
V_v = W_V^v F_v
  \label{eq:important}
\end{equation}
Above vectors are fed into a multi-head attention module for fusion to compute a text-guided refinement of the visual representation, as in Eq. (4).

\begin{equation}
F_{v \leftarrow t} = \mathrm{MHA}(Q_t, K_v, V_v)
  \label{eq:important}
\end{equation}
Then the output is passed through a residual connection with the original query and followed by LayerNorm, as in Eq. (5). 
\begin{equation}
F_v' = \mathrm{LayerNorm}(F_t + F_{v \leftarrow t})
  \label{eq:important}
\end{equation}
To further enhance the representational capacity of the fused features, we introduce a Gated Feed-Forward Network (GatedFFN) on top of the cross-modal attention output $F_v'$. Inspired by the gated feed-forward modules in Transformer architectures\cite{han2022survey}, GatedFFN incorporates non-linear transformations to enable deeper intra-modal modeling after cross-modal fusion. The final refined visual feature is obtained via another residual connection followed by LayerNorm, as in Eq. (6).
\begin{equation}
\label{eq:final_refined_visual}
F_v^{r} = \mathrm{LayerNorm}(F_{v}^{'} +\mathrm{GatedFFN}( F_{v}^{'}))
\end{equation}
A symmetric operation is performed for vision-guided text refinement, resulting in $F_t^{r}$. The refined features $F_v^{r}$ and  $F_t^{r}$ are used in subsequent projection and retrieval computation. 

This design addresses a key limitation in prior CLIP-like dual encoders: the lack of contextual alignment between modalities before computing similarity. Our CMCR enables each modality to absorb semantic cues from the other, allowing for improved alignment between, for example, specific actions in a video and verbs in a query sentence, or object compositions in cluttered scenes and textual descriptions.
%-------------------------------------------------------------------------
\subsection{Loss Functions}

The refined visual and textual features are mapped to a common $d$  embedding space by nonlinear projection (two layers of MLP with GELU activation function) and normalized to obtain the final visual embedding and textual embedding.

The model is trained using Symmetric Multi-Similarity Loss (SMSLoss). This loss function is particularly suitable for multi-instance retrieval tasks with soft labels, such as correlation matrix $C$ in EPIC-KITCHENS-100. It models both visual-to-text and text-to-visual directions symmetrically and redefines the relationship between positive and negative sample pairs through the correlation matrix $C_{ij}$, as in Eq. (7). 
\begin{equation}
R = C_{ij} - C_{ik}
  \label{eq:important}
\end{equation}
In addition, a relaxation factor $\tau$ is introduced to mitigate the detrimental effect of samples with minimal similarity differences on the training, improving the model's ability to model soft-label relationships.

%-------------------------------------------------------------------------
\subsection{Test-Time Augmentation}

To enhance the robustness of feature representations during inference, we extend the test-time augmentation (TTA) strategy initially adopted in SMSLoss by introducing multi-scale augmentation as a complementary component. While SMSLoss proposes an effective similarity learning objective and employs horizontal flipping at test time to capture viewpoint variations, its augmentation scheme remains relatively simple. This limits its ability to handle spatial scale variations, which can lead to suboptimal generalization in complex video scenarios. 

To address this limitation, we propose an enhanced TTA strategy that combines both horizontal flipping and multi-scale augmentation. Specifically, the model first extracts features from the original input as the base representation. If flipping is enabled, video frames are horizontally fliped and re-encoded. In parallel, multi-scale augmentation resizes the video frames to 0.875×, 1.0× and 1.125× of the original resolution, followed by center cropping to restore the original size and feature extraction at each scale. 

All  feature variants are aggregated via average pooling to produce a single, context-aware, and semantically robust representation for subsequent similarity computation. This strategy significantly improves the model's ability to handle variations in orientation and spatial scale during inference, while maintaining efficiency and strong generalization. 
\section{Experimemts}

\subsection{Experimental Details }

We directly build upon the pretrained model from SMSLoss and AVION, which is a vanilla CLIP-based architecture\cite{radford2021learning} trained on the LLM-augmented Ego4D dataset\cite{grauman2022ego4d}\cite{zhao2023learning}. We then fine-tune our ContextRefine-CLIP on the EPIC-KITCHENS-100 dataset.

All experiments are conducted on 4× NVIDIA GeForce RTX 3090 GPUs, using the same training setup as SMSLoss. For our ViT-L based model, each 24GB GPU fits 40 video clips, resulting in a total batch size of 160. The learning rate is set to 1.8e-5. For our cross-modal refinement module, we use an attention of 512, 8 attention heads, and dropout rate of 0.1 across all components.

\subsection{Results and Analysis}

Due to time constraints, we explored directly based on ViTL-14. Consistent with SMSLoss, we retained the first three valid digits in the results without rounding, and all the experiments were conducted with the same learning rate and optimizer settings. Table 1 shows the detailed comparison results of our proposed CR-CLIP model and its key components, covering the two main metrics, mAP and nDCG, which respectively measure retrieval accuracy and ranking relevance.

Compared to the traditional MI-MM loss, SMSLoss achieves notable performance gains, improving the average mAP by 6.4\% and the average nDCG by 2.4\%, thereby validating its effectiveness for multi-instance retrieval. Building upon SMSLoss, our proposed CR-CLIP incorporates the CMCR module, which further enhances performance without GatedFFN or test-time augmentation. Specifically, CR-CLIP achieves an average mAP of 66.2\% and an average nDCG of 80.9\%, surpassing SMSLoss by 4.1\% and 7.9\%, respectively. These results highlight the strong effectiveness of the CMCR module in improving semantic alignment and feature discriminability through cross-modal contextual interactions, with particularly significant gains observed in the nDCG metric.

After introducing the GatedFFN component on top of the CMCR module (w/ GatedFFN), the T2V mAP improves from 60.5\% to 61.1\%, and the average nDCG increases from 80.9\% to 81.6\%. Although a slight drop is observed in the V2T mAP, the overall ranking performance improves, indicating that GatedFFN effectively enhances the representational capacity of the fused features. Building upon this, incorporating horizontal flip augmentation (w/ Flip) further improves the average mAP and nDCG to 66.6\% and 81.9\%, respectively. When combined with multi-scale augmentation (w/ Flip+Scale), the model achieves the best performance, with the average mAP and nDCG reaching 66.8\% and 82.1\%, respectively. These results demonstrate that multi-scale augmentation plays a crucial role in handling spatial scale variations and complements horizontal flipping, leading to enhanced robustness and generalization. 

\begin{table}[t]
\centering
\begin{tabular}{ccccccc}
\toprule
 {\textbf{Methods}}& \multicolumn{3}{c}{\textbf{mAP (\%)}} & \multicolumn{3}{c}{\textbf{nDCG (\%)}} \\
 & V2T& T2V& Avg.& V2T& T2V& Avg. \\
\midrule
 MI-MM & 58.7 & 52.7 & 55.7 & 71.9 & 69.4 & 70.6 \\
 SMSLoss& 67.3 & 56.9& 62.1 & 74.7& 71.2& 73.0 \\
 CR-CLIP& 71.8& 60.5& 66.2& 82.5& 79.2&80.9\\
 w/GatedFFN& 71.2& 61.1& 66.2& 83.3& 79.8&81.6\\
 w/ Flip& 71.6& 61.5& 66.6& 83.7& 80.1&81.9\\
  w/Flip+Scale& 71.8& 61.8& 66.8& 83.9& 80.3& 82.1\\
\bottomrule
 & & & & & &\\
\end{tabular}
  \caption{ comparison of the CR-CLIP  and test-time augmentation strategies on the EK-100 dataset.}
  \label{tab:main_results}
\end{table}

\section{Conclusion}

In this report, we focus on enhancing vision-language multi-instance retrieval, particularly in complex scenarios such as EPIC-KITCHENS-100 where soft labels are provided via relevance matrices. We propose ContextRefine-CLIP, a dual-encoder architecture that introduces Cross-Modal Context Refinement. By enabling deep bidirectional interactions through cross-attention and  Gated Feed-Forward Network, CMCR significantly improves contextual understanding and semantic alignment between modalities. When combined with the Symmetric Multi-Similarity Loss, CR-CLIP achieves superior performance, reaching 66.8\% average mAP and 82.1\% average nDCG without ensembling, outperforming existing baselines. Additionally, we demonstrate that test-time augmentation strategies—specifically horizontal flipping and multi-scale inference—further enhance model robustness and retrieval accuracy. 

Despite the promising results, several limitations remain. First, we do not explore model ensembling due to time and resource constraints, integrating multiple CR-CLIP variants or combining with other models may further boost performance. Second, while the CMCR module is designed to be lightweight, we lack a detailed analysis of its computational overhead and parameter efficiency. Third, our experiments focus primarily on ViT-L, testing on smaller backbones like ViT-B would help assess scalability and generalizability. Lastly, the hyperparameter tuning space for CMCR is limited and broader exploration may lead to even better configurations. 
{
    \small
    \bibliographystyle{ieeenat_fullname}
    \bibliography{main}

\begin{thebibliography}{11}
\providecommand{\natexlab}[1]{#1}
\providecommand{\url}[1]{\texttt{#1}}
\expandafter\ifx\csname urlstyle\endcsname\relax
  \providecommand{\doi}[1]{doi: #1}\else
  \providecommand{\doi}{doi: \begingroup \urlstyle{rm}\Url}\fi

\bibitem[Damen et~al.(2018)Damen, Doughty, Farinella, Fidler, Furnari, Kazakos, Moltisanti, Munro, Perrett, Price, et~al.]{damen2018scaling}
Dima Damen, Hazel Doughty, Giovanni~Maria Farinella, Sanja Fidler, Antonino Furnari, Evangelos Kazakos, Davide Moltisanti, Jonathan Munro, Toby Perrett, Will Price, et~al.
\newblock Scaling egocentric vision: The epic-kitchens dataset.
\newblock In \emph{Proceedings of the European conference on computer vision (ECCV)}, pages 720--736, 2018.

\bibitem[Damen et~al.(2022)Damen, Doughty, Farinella, Furnari, Kazakos, Ma, Moltisanti, Munro, Perrett, Price, et~al.]{damen2022rescaling}
Dima Damen, Hazel Doughty, Giovanni~Maria Farinella, Antonino Furnari, Evangelos Kazakos, Jian Ma, Davide Moltisanti, Jonathan Munro, Toby Perrett, Will Price, et~al.
\newblock Rescaling egocentric vision: Collection, pipeline and challenges for epic-kitchens-100.
\newblock \emph{International Journal of Computer Vision}, pages 1--23, 2022.

\bibitem[Grauman et~al.(2022)Grauman, Westbury, Byrne, Chavis, Furnari, Girdhar, Hamburger, Jiang, Liu, Liu, et~al.]{grauman2022ego4d}
Kristen Grauman, Andrew Westbury, Eugene Byrne, Zachary Chavis, Antonino Furnari, Rohit Girdhar, Jackson Hamburger, Hao Jiang, Miao Liu, Xingyu Liu, et~al.
\newblock Ego4d: Around the world in 3,000 hours of egocentric video.
\newblock In \emph{Proceedings of the IEEE/CVF conference on computer vision and pattern recognition}, pages 18995--19012, 2022.

\bibitem[Han et~al.(2022)Han, Wang, Chen, Chen, Guo, Liu, Tang, Xiao, Xu, Xu, et~al.]{han2022survey}
Kai Han, Yunhe Wang, Hanting Chen, Xinghao Chen, Jianyuan Guo, Zhenhua Liu, Yehui Tang, An Xiao, Chunjing Xu, Yixing Xu, et~al.
\newblock A survey on vision transformer.
\newblock \emph{IEEE transactions on pattern analysis and machine intelligence}, 45\penalty0 (1):\penalty0 87--110, 2022.

\bibitem[Le-Khac et~al.(2020)Le-Khac, Healy, and Smeaton]{le2020contrastive}
Phuc~H Le-Khac, Graham Healy, and Alan~F Smeaton.
\newblock Contrastive representation learning: A framework and review.
\newblock \emph{Ieee Access}, 8:\penalty0 193907--193934, 2020.

\bibitem[Li et~al.(2019)Li, Tao, Li, and Fu]{li2019visual}
Sheng Li, Zhiqiang Tao, Kang Li, and Yun Fu.
\newblock Visual to text: Survey of image and video captioning.
\newblock \emph{IEEE Transactions on Emerging Topics in Computational Intelligence}, 3\penalty0 (4):\penalty0 297--312, 2019.

\bibitem[Radford et~al.(2021)Radford, Kim, Hallacy, Ramesh, Goh, Agarwal, Sastry, Askell, Mishkin, Clark, et~al.]{radford2021learning}
Alec Radford, Jong~Wook Kim, Chris Hallacy, Aditya Ramesh, Gabriel Goh, Sandhini Agarwal, Girish Sastry, Amanda Askell, Pamela Mishkin, Jack Clark, et~al.
\newblock Learning transferable visual models from natural language supervision.
\newblock In \emph{International conference on machine learning}, pages 8748--8763. PmLR, 2021.

\bibitem[Wang et~al.(2024{\natexlab{a}})Wang, Wang, and Chau]{wang2024sms}
Xiaoqi Wang, Yi Wang, and Lap-Pui Chau.
\newblock Symmetric multi-similarity loss for epic-kitchens-100 multi-instance retrieval challenge 2024, 2024{\natexlab{a}}.

\bibitem[Wang et~al.(2024{\natexlab{b}})Wang, Chen, Yan, Jamieson, and Du]{wang2024variance}
Yiping Wang, Yifang Chen, Wendan Yan, Kevin Jamieson, and Simon~Shaolei Du.
\newblock Variance alignment score: A simple but tough-to-beat data selection method for multimodal contrastive learning.
\newblock \emph{arXiv preprint arXiv:2402.02055}, 2024{\natexlab{b}}.

\bibitem[Zhao and Kr{\"a}henb{\"u}hl(2023)]{zhao2023training}
Yue Zhao and Philipp Kr{\"a}henb{\"u}hl.
\newblock Training a large video model on a single machine in a day, 2023.

\bibitem[Zhao et~al.(2023)Zhao, Misra, Kr{\"a}henb{\"u}hl, and Girdhar]{zhao2023learning}
Yue Zhao, Ishan Misra, Philipp Kr{\"a}henb{\"u}hl, and Rohit Girdhar.
\newblock Learning video representations from large language models.
\newblock In \emph{Proceedings of the IEEE/CVF Conference on Computer Vision and Pattern Recognition}, pages 6586--6597, 2023.

\end{thebibliography}
}


\end{document}